\documentclass[letterpaper, 10 pt, conference]{ieeeconf}

\input{sections/0_preamble}
\title{\huge \bf
An Interleaved Approach to Trait-Based Task Allocation and Scheduling
}
\author{Glen Neville*$^\dagger$,
Andrew Messing*$^\dagger$,
Harish Ravichandar$^\dagger$,
Seth Hutchinson$^\dagger$ 
and Sonia Chernova$^\dagger$ %<-this % stops a space

\thanks{This work was supported by the Army Research Lab under Grants W911NF-17-2-0181 (DCIST CRA) and W911NF-20-2-0036}% <-this % stops a space

\thanks{$^\dagger$Georgia Institute of Technology {\tt\small \{gneville, amessing, harish.ravichandar, seth, chernova\}@gatech.edu}}%
\thanks{$\textsuperscript{*}$Authors contributed equally to this research.}%
}

\newcommand{\possibleBold}[2]{%
    \ensuremath{%
        \ifthenelse{\equal{#1}{bold}}%
        {\boldsymbol{#2}}%
        {#2}%
    }%
}
\newcommand{\taskNetwork}[1][none]{\possibleBold{#1}{\mathcal{T}}}
\newcommand{\robotTraitMatrix}[1][none]{\possibleBold{#1}{Q}}
\newcommand{\desiredTraitMatrix}[1][none]{\possibleBold{#1}{Y^*}}
\newcommand{\numRobots}[1][none]{\possibleBold{#1}{N}}
\newcommand{\numTasks}[1][none]{\possibleBold{#1}{M}}

\newcommand{\problemDomain}[1][none]{\possibleBold{#1}{\mathcal{D}}}

\newcommand{\configSpace}[1][none]{\possibleBold{#1}{\mathcal{C}}}

\newcommand{\setInitialConfigs}[1][none]{\possibleBold{#1}{\boldsymbol{I_c}}}
\newcommand{\setConfigSpaces}[1][none]{\possibleBold{#1}{\mathcal{X}}}
\newcommand{\allocation}[1][none]{\possibleBold{#1}{\mathcal{A}}}

\newcommand{\schedule}[1][none]{\possibleBold{#1}{\sigma}}
\newcommand{\setMotionPlans}[1][none]{\possibleBold{#1}{\mathcal{M}}}

\begin{document}

\maketitle
\thispagestyle{empty}
\pagestyle{empty}

\begin{abstract}

% To effectively realize heterogeneous multi-robot teams, we must leverage individual robots' relative strengths as well as respect how their motions and assignments affect the overall mission completion time. 
To realize effective heterogeneous multi-robot teams, researchers must leverage individual robots' relative strengths and coordinate their individual behaviors. Specifically, heterogeneous multi-robot systems must answer three important questions: \textit{who} (task allocation), \textit{when} (scheduling), and \textit{how} (motion planning). While specific variants of each of these problems are known to be NP-Hard, their interdependence only exacerbates the challenges involved in solving them together. In this paper, we present a novel framework that interleaves task allocation, scheduling, and motion planning. We introduce a search-based approach for trait-based time-extended task allocation named Incremental Task Allocation Graph Search (ITAGS). In contrast to approaches that solve the three problems in sequence, ITAGS's interleaved approach enables efficient search for allocations while simultaneously satisfying scheduling constraints and accounting for the time taken to execute motion plans. To enable effective interleaving, we develop a convex combination of two search heuristics that optimizes the satisfaction of task requirements as well as the makespan of the associated schedule. We demonstrate the efficacy of ITAGS using detailed ablation studies and comparisons against two state-of-the-art algorithms in a simulated emergency response domain.

\end{abstract}

\section{Introduction}

% Heterogenous multi-robot systems are cool but hard
Heterogeneous multi-robot systems offer the potential to solve large-scale problems that involve multiple interdependent tasks and require a diverse set of capabilities. As such, multi-robot systems (MRSs) have proved useful in several domains, including agriculture\cite{Roldan2016HeterogeneousGreenhouses}, military\cite{McCook2007a}, assembly\cite{Stroupe2005HeterogeneousServicing}, and warehouse automation\cite{Baras2019AnWarehouses}.
However, it is not trivial to efficiently solve complex MRS problems, e.g. coalition formation, due to the combinatorial complexities of coordinating a team of heterogeneous robots. Indeed, problems associated with optimally allocating heterogeneous robots to tasks are known to be strongly NP-hard~\cite{Gerkey2004ASystems,Korsah2013}.

% We are solving the hard problem of ST-MR-TA (coalition formation + scheduling problem) along with motion planning
In this work, we present an approach to simultaneously address coalition formation and scheduling for heterogeneous multi-robot systems. The coalition formation problem is concerned with identifying which subset of robots should perform each task, and the scheduling problem involves identifying when each task should be performed. As such, we solve a particularly challenging variant of the multi-robot task allocation (MRTA) problem targeting single-task robots, multi-robot tasks, and time-extended allocation (ST-MR-TA)~\cite{Gerkey2004ASystems,Korsah2013}. Further, our approach can account for inter-task dependencies (expressed in the form of ordering and non-concurrency constraints) and satisfy temporal constraints imposed by motion planning.

% We have a novel framework that interleaves three modules
We contribute a novel framework consisting of three interconnected modules -- task allocation, scheduling, and motion planning. A key attribute of our framework is the effective information exchange among the different modules (see \Cref{fig:pipeline}). In contrast to solving these problems in sequence, our interleaved approach enables efficient generation of allocations, schedules, and motion plans that do not conflict with each other. We demonstrate that our interleaved approach improves allocation quality, scheduling efficiency, and overall computational efficiency.

% We make specific contributions to heterogeneous task allocation
In addition, we make specific contributions to task allocation for heterogeneous multi-robot teams. Specifically, we introduce a search-based approach to task allocation, named Incremental Task Allocation Graph Search (ITAGS). ITAGS computes an allocation of robots to tasks while \textit{simultaneously} optimizing the satisfaction of task requirements and the makespan (i.e. execution time) of the associated schedule. 
% Our framework is compatible with existing approaches to scheduling and motion planning in the sense that .  

% trait-based task allocation and search
We leverage recent advances in instantaneous task allocation~\cite{Ravichandar2019,Prorok2017TheSwarms}, and model our task requirements in terms of the traits (i.e. capabilities) required for each task. Such trait-based specifications do not require the user to explicitly specify the relationships between each task and robot, allowing for generalization to different types of robots. 

% The two heuristics
We enable effective interleaving of task allocation, scheduling, and motion planning by introducing a convex combination of two heuristics in our iterative search algorithm. The first heuristic measures how well the current allocation satisfies the specified trait requirements. The second heuristic measures the efficiency of the schedule associated with the current allocation. We note that our second heuristic accounts for both the feasibility and duration of all motion plans when evaluating a schedule.

%Intuition for heuristics and implications
Combined, our two heuristics allow ITAGS to prune branches of the incremental task allocation graph that are infeasible to execute due to constraint violations either in scheduling or motion planning. Such pruning allows ITAGS to prefer satisficing incremental allocations, resulting in significant improvements in overall computational efficiency and solution quality. Further, we provide practical insights into the relative influence of each of the two heuristics on ITAG's performance. 

\newpage

In summary, we contribute:

\begin{enumerate}
    \item a unified framework that interleaves task allocation, scheduling, and motion planning for heterogeneous multi-robot systems,
    \item a novel graph-based search algorithm for trait-based task allocation, and
    \item two complementary heuristic functions that help efficiently search the task allocation graph while accounting for scheduling and motion planning constraints.
\end{enumerate}

To illustrate the impact of our contributions to the robotics community, we evaluate our framework using experiments in a simulated emergency response domain. Firstly, our experiments systematically investigate both individual and combined effects of the two heuristics in terms of convergence, computational cost, makespan, and search efficiency. Secondly, we compare the proposed framework against a baseline approach that does not interleave allocation, scheduling, and motion planning. Thirdly, we compare the proposed framework against two recent state-of-the-art task allocation algorithms. The results of these evaluations conclusively demonstrate the efficacy and necessity of our framework.

\section{Related Works}

A rich body of work has addressed the multi-robot task allocation (MRTA) problem~\cite{Gerkey2003Multi-robotArchitectures,Korsah2013} and the closely-related scheduling problem \cite{Nunes2017}. Our work addresses a variant of the MRTA problem that involves single-task (ST) robots, multi-robot (MR) tasks, and time-extended allocation (TA). As such, we limit our discussion of related work to methods that address the ST-MR-TA variant of MRTA.

% Auction-based methods
A popular approach to solve the ST-MR-TA problem has been the use of auction-based methods in which robots and tasks are matched using a bidding process based on models of how well each robot can perform each task\cite{Sariel2006, Jones2011Time-extendedConstraints, Giordani2013, Krizmancic2020}.
However, these methods require that each multi-robot task be decomposed into multiple single-robot tasks or that the user specifies the distribution of robots for each task. Additionally, they either schedule tasks then allocate them or vise versa. ITAGS, on the other hand, does not require decomposition of tasks nor a specification on the number of robots needed for each task, and it interleaves task allocation and scheduling, which allows it to create more efficient schedules than possible when operating sequentially.

% Optimization-based methods
Optimization methods, while primarily used for solving scheduling and ST-SR-TA variant of the MRTA problem (see \cite{Gerkey2003Multi-robotArchitectures} for more information on the ST-SR-TA problem), have also been utilized to solve the ST-MR-TA problem \cite{Tompkins2003OptimizationOperationsb, Korsah2012, Guerrero2017Multi-robotSolutions}. 
Optimization-based approaches typically cast task allocation as a mixed-integer linear program. However, each of these approaches either does not require all tasks to be completed or requires decomposition of multi-robot tasks. In comparison, ITAGS requires that all tasks be accomplished and does not require the decomposition of tasks.

% Tree/Graph-based methods  
Recent efforts have formulated the ST-MR-TA problem as tree and graph-search problems \cite{Thakar2019TaskManipulation, Kabir2020IncorporatingManipulators}. 
Both of the approaches in \cite{Thakar2019TaskManipulation} and \cite{Kabir2020IncorporatingManipulators} first schedule tasks into a sequence of temporal windows and then allocate robots to these tasks. ITAGS also conducts a tree search, however, does not use temporal windows, which allows for a higher amount of concurrency between tasks, and simultaneously considers both scheduling and allocation through its heuristics.

Some authors have proposed using processor scheduling techniques to solve the ST-MR-TA problem \cite{Ramchurn2010CoalitionConstraints, Capezzuto2020}. Recently, Capezzuto \etal\cite{Capezzuto2020} proposed Coalition Formation with Improved Look-Ahead (CFLA2) and Cluster-based Coalition Formation (CCF), two algorithms based on processor scheduling techniques, to solve what they call the Coalition Formation with Spatial and Temporal Constraints Problem (CFSTP). These approaches do not require all tasks to be completed when they allocate. On the other hand, ITAGS ensures that all tasks are executed.

While the methods discussed thus far have numerous advantages, a common limitation is that they assume that the desired behavior is specified in terms of an optimal robot distribution or a utility function describing how well each robot can perform each task. In contrast, recent advances have enabled modeling of task requirements in terms of desired trait distributions~\cite{Koes2005, Prorok2017TheSwarms, Ravichandar2019}.
These trait-based models are more robust to changes in the number of robots. However, the existing approaches presented in \cite{Koes2005, Prorok2017TheSwarms} are limited to binary traits. These binary traits are less expressive than continuous traits and lead to less robust models of the agents in a multi-agent team. In addition, the approaches presented in \cite{Prorok2017TheSwarms, Ravichandar2019} are limited to the ST-MR-IA (instantaneous allocation) problem. They do not consider scheduling, making these solutions unable to handle the time-extended domain of our problems. We take inspiration from these methods and propose an approach for continuous trait-based ST-MR-TA that simultaneously solves task allocation, scheduling, and motion planning.

%In general, existing approaches do not iteratively share information between the component parts of time-extended task allocation (scheduling, task allocation, motion planning) effecting the quality of the solutions found. We use a trait-based model, which allows our method to model the robot of large heterogeneous teams better and pair it with a novel encoding of task allocation and heuristics that allow for trait-based time-extended task allocation. 

\section{Problem Description}
Assigning tasks to the different robots and coalitions requires reasoning about their complementary traits and the team's limited resources. These assignments must also respect robot motion planning and scheduling. In this section, we discuss the formulation of a trait-based time-extended allocation problem in two parts. First, we present the elements of the problem domain. Second, we describe the format of a solution in this domain.

\subsection{Problem Domain}
%% Agent Trait Vectors
Consider a heterogeneous team of \numRobots[bold]\ robots, where each robot is defined by its abilities or \textit{traits}, which are modeled as continuous variables. Each robot's traits are defined as
$$
    q^{(i)} = \sqBrackets{q_1^{(i)},\ q_2^{(i)},\ \cdots, q_U^{(i)}}
$$
where $q_u^{(i)} \in \R_{+}$ corresponds to the $u^{th}$ trait for the $i^{th}$ robot. If the $i^{th}$ robot does not have the $u^{th}$ trait (e.g. firetrucks have a water capacity, but other robots may not) then $q_u^{(i)} = 0$, otherwise it is a positive value. The traits of the entire team are defined by the \textbf{robot trait matrix} %\robotTraitMatrix[bold]\ where
$$
  \robotTraitMatrix[bold]  = \sqBrackets{q^{(1)^{\intercal}},\ \cdots,\ q^{(N)^{\intercal}}}^{\intercal} \in \R_{+}^{N \times U}
$$
with each row corresponding to one robot and each column corresponding to one trait. 

A \textbf{Task Network} is a directed graph $G=(\mathcal{E}, \mathcal{V})$. The vertices $\mathcal{V}$ represent a set of tasks. The edges $\mathcal{E}$ connect tasks such that an edge $e = [t_i, t_j] \quad t_i, t_j \in \mathcal{V}$ represents a \textbf{precedence constraint} ($t_i \prec t_j$), or a relationship that ensures that $t_i$ concludes before $t_j$ starts \cite{Weld1994AnPlanning}.

% %% Precedence Constraint
% \begin{definition}
% \label{def::precedence_constraint}
% A \textbf{precedence constraint} ($t_a \prec t_b$) is a relationship between two tasks, $t_a$ and $t_b$, that ensures that execution of $t_a$ concludes before $t_b$ starts\cite{Weld1994AnPlanning}.
% \end{definition}

% %% Task Network
% \begin{definition}
% \label{def:task_network}
% A \textbf{Task Network} is a directed graph $G=(\mathcal{E}, \mathcal{V})$. The vertices $\mathcal{V}$ represent a set of tasks and the arcs $\mathcal{E}$ connect tasks such that an arc $e = [v_i, v_j] \quad v_i, v_j \in \mathcal{V}$ represents a precedence constraint $v_i \prec v_j$.
% \end{definition}

% The topology of the tasks is modeled using a \textbf{task network} $\boldsymbol{\mathcal{T}}$. 
Given the robot trait matrix \robotTraitMatrix[bold], the team is required to accomplish \numTasks[bold]\ tasks from a task network \taskNetwork[bold]. Each task in \taskNetwork[bold]\ is defined by the traits required to accomplish it, a static duration, and its initial and terminal configuration (e.g., to move a box, one or more robots need to be at the box's location to start the task and will be at the terminal location of the move upon finishing it). Robots can either complete tasks individually or collaborate on tasks as part of a coalition, depending on their traits. The desired traits for a task are defined as follows:
$$
    y^{(i)} = \sqBrackets{y_1^{(i)}, y_2^{(i)}, \cdots, y_U^{(i)}}
$$
where $y_u^{(i)} \in \R_{+}$ is the $u^{th}$ trait requirement for the $i^{th}$ task. If the $u^{th}$ trait is not required by the $i^{th}$ task then $y_u^{(i)} = 0$, otherwise it is a positive value. The tasks required by the entire task network is defined by the \textbf{desired trait matrix} %\desiredTraitMatrix[bold]\ where
$$
  \desiredTraitMatrix[bold]  = \sqBrackets{y^{(1)^{\intercal}},\ \cdots,\ y^{(N)^{\intercal}}}^{\intercal} \in \R_{+}^{M \times U}
$$
with each row corresponding to one task and each column corresponding to one trait.

For a robot to participate in completing a task, it needs a collision-free path through its configuration space \configSpace[bold]\cite{Choset2005PrinciplesImplementations} from its current configuration to the initial configuration of the task, and then it needs a collision-free path to the terminal configuration of the task. Each type of robot can have a different free configuration space (e.g. a quadcopter can fly over obstacles that a ground vehicle cannot).

%We define the domain for this problem as the tuple $\problemDomain[bold] = \angled{\taskNetwork[bold],\ \robotTraitMatrix[bold],\ \desiredTraitMatrix[bold],\ \setInitialConfigs[bold],\ \setConfigSpaces[bold]}$ where
% \begin{flushitemize}
%     \item \taskNetwork[bold]\ is a task network
%     \item \robotTraitMatrix[bold]\ is the robot trait matrix
%     \item \desiredTraitMatrix[bold]\ is the desired trait matrix
%     \item \setInitialConfigs[bold]\ is a finite set of the initial configurations for each robot
%     \item \setConfigSpaces[bold]\ is a finite set of free configuration spaces for each type of robot
% \end{flushitemize}

%% Non-list version
We define the domain as the tuple $\problemDomain[bold] = \angled{\taskNetwork[bold],\ \robotTraitMatrix[bold],\ \desiredTraitMatrix[bold],\ \setInitialConfigs[bold],\ \setConfigSpaces[bold]}$ where
\taskNetwork[bold]\ is a task network, \robotTraitMatrix[bold]\ is the robot trait matrix, \desiredTraitMatrix[bold]\ is the desired trait matrix,  \setInitialConfigs[bold]\ is a set of the initial configurations with one for each robot, and \setConfigSpaces[bold]\ is a set of free configuration spaces with one for each type of robot.

\subsection{Solution Specification}
Given the above problem domain, we now introduce the various parts of the solution.

An \textbf{allocation} \allocation[bold]\ for 
%a robot trait matrix $Q$ with 
\numRobots\ robots and 
%a task network $\mathcal{T}$ with 
\numTasks\ tasks is a $\numTasks \times \numRobots$\ matrix  where 
\[ \allocation_m^n = \begin{cases}
     1 & \text{if the $n^{th}$ robot is assigned to the $m^{th}$ task.}\\
    0 & \textit{otherwise.}
 \end{cases}
\]

Given the problem domain $\boldsymbol{\mathcal{D}}$, we wish to compute the \textbf{solution} $\boldsymbol{\mathcal{S} = \angled{\allocation,\ \setMotionPlans,\ \schedule}}$ where \allocation[bold]\ is an allocation that satisfies the desired traits of \taskNetwork[bold], \setMotionPlans[bold]\ is a finite set of motion plans containing the motion plans for each robot to arrive at the initial configuration for each task it is assigned to complete as well as the motion plans needed for completing each task in \taskNetwork[bold], and \schedule[bold]\ is a schedule with the minimum makespan. The schedule \schedule[bold] also includes the time to execute the motion plans in \setMotionPlans[bold]\ and is temporally consistent (i.e. respects all temporal constraints, such as ordering and mutex constraints).

\section{Approach}
%Our method is part of a larger hierarchical framework that includes layers for symbolic task planning and motion planning; however, for this paper, we focus on the method used by the task allocation and scheduling layers (see \Cref{fig:pipeline}). 
This section outlines the high-level algorithmic architecture that we use for trait-based time-extended task allocation for groups of heterogeneous robots. After introducing the high-level architecture, the following subsections will explain individual layers of the hierarchy. 

%The task network \taskNetwork[bold]\ and desired trait matrix \desiredTraitMatrix[bold]\ are provided by a task planning layer; however, as that layer is not a focus in this paper, it will be assumed that those are provided as part of the problem. 
It is assumed that the task network \taskNetwork[bold], desired trait matrix \desiredTraitMatrix[bold], the robot trait matrix \robotTraitMatrix[bold], a set of the initial configurations \setInitialConfigs[bold], and a set of configuration spaces for each type of robot \setConfigSpaces[bold] are provided as presented in the previous section.

\begin{figure}[t]
\includegraphics[width=1.0\columnwidth, trim=0 5 0 0, clip]{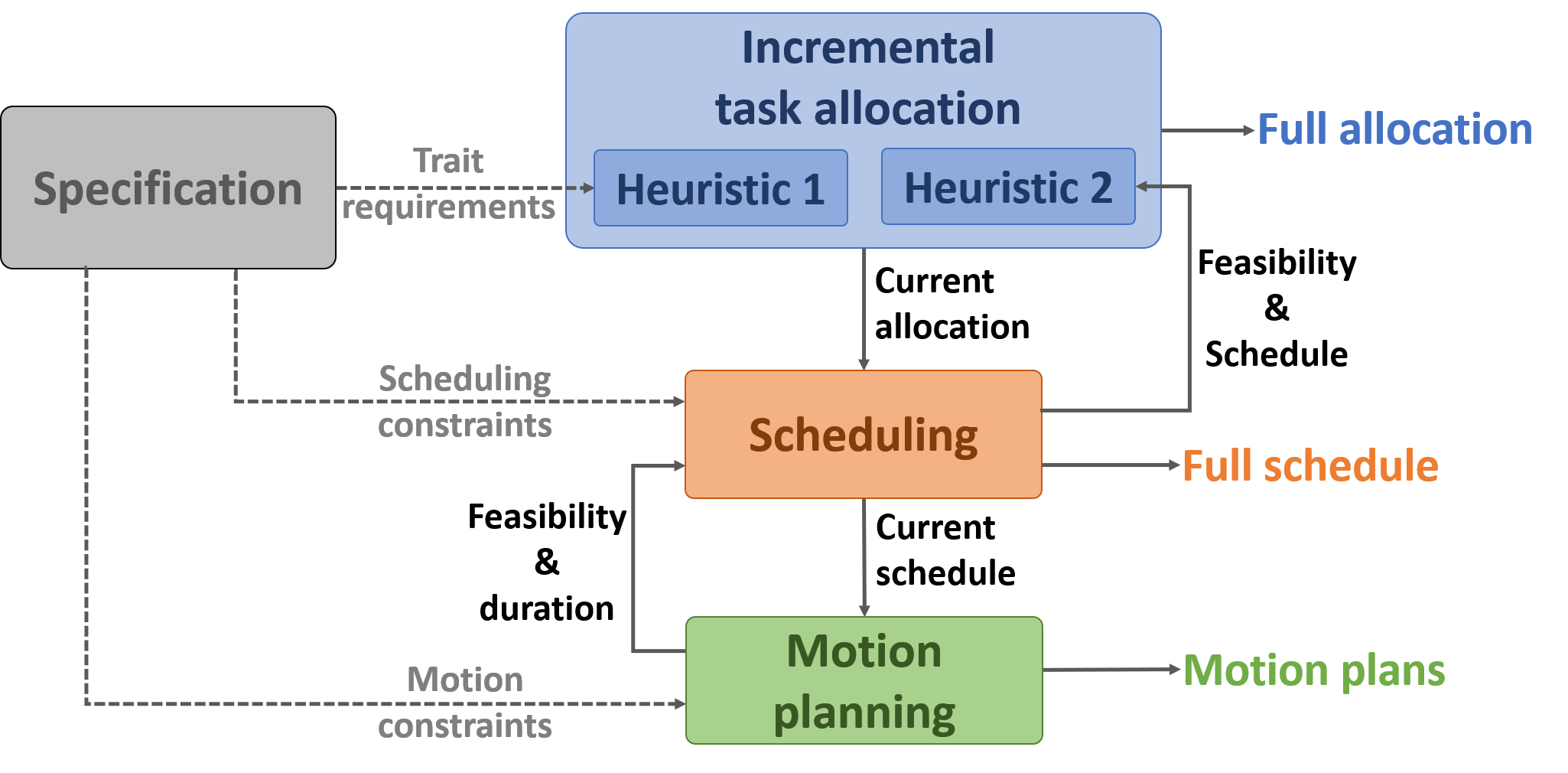}
\caption{High-level architecture of the hierarchical framework.}
\label{fig:pipeline} 
\end{figure}

The task allocation layer conducts a heuristic graph search through an incremental task allocation space to find an allocation that satisfies the desired trait requirements of the task network. During the search, it provides allocations to the scheduling layer, which informs the heuristic that guides its search and allows it to prune areas of the incremental task allocation space that are not feasible to schedule. The scheduling layer computes the schedule with the minimum makespan for each allocation. The motion planning layer uses a generic motion planner to determine if there is a feasible motion plan between two configurations for a robot or coalition based on provided configuration spaces as well as the execution time for the motion plan.

This hierarchical method iterates between the layers until it finds an allocation that satisfies the trait requirements of the task network and is feasible to schedule while respecting the execution times for the motion plans (see Figure \ref{fig:pipeline}). It then outputs the task allocation \allocation[bold], the associated schedule \schedule[bold], and required motion plans \setMotionPlans[bold]\ needed for the schedule. Algorithm \ref{algo:itags} contains the pseudo-code for ITAGS.

\subsection{Task Allocation} 
\begin{figure}[t]
\begin{center}
    \includegraphics[width=0.9\columnwidth]{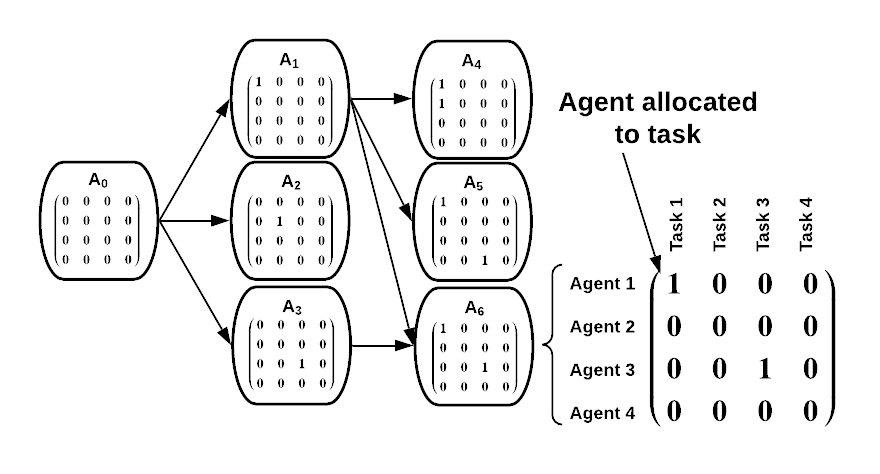}
\end{center}
\caption{An example incremental task allocation graph. %with three robots and three tasks
}
\label{fig:alloc} 
\end{figure}

The task allocation layer performs a greedy best-first search through the incremental task allocation space. In this space, nodes represent an allocation of robots to tasks. Nodes are connected to other nodes that differ only by the assignment of a single robot (see \Cref{fig:alloc}). This graphical representation allows our search to start from an initial node with no allocated robots and to incrementally add robots until an allocation that satisfies the desired trait requirements of the task network is found. 

To guide the search, we have developed two heuristics: \textit{Allocation Percentage Remaining}, which guides the search based on the quality of the allocation, and \textit{Normalized Schedule Quality}, which guides the search based on the quality of the makespan of the schedule associated with the allocation. We use a convex combination of the two heuristics, which we call \textit{Time-Extended Task Allocation Quality}.

\subsubsection{Allocation Percentage Remaining (APR)}
APR is defined as the percentage trait mismatch error. Specifically, APR is calculated as
% \begin{equation}
%     \label{equ:mismatch_error}
%     E(\overline{\allocation}) = \desiredTraitMatrix - \overline{\allocation}\ Q
% \end{equation}
\begin{equation}
    \label{equ:apr}
    f_{apr}(\overline{\allocation}) = \frac{||\max(E(\overline{\allocation}),\ 0)||_{1,1}}{ ||\desiredTraitMatrix||_{1,1}}
\end{equation}
where $\overline{\allocation}$ is the allocation that the heuristic is evaluating, $E(\overline{\allocation}) = \desiredTraitMatrix - \overline{\allocation}\ Q$, and $||\cdot||_{1,1}$ is the element-wise $l_1$ norm. In this equation, $(\overline{A} Q)_{i, j}$ is the summation of the $j^{th}$ trait from each of the robots assigned to the $i^{th}$ task. By taking the difference between the desired trait matrix \desiredTraitMatrix\ and $\overline{A} Q$, we get the \textit{trait mismatch matrix} $E$. When $E_{i, j} < 0$ then the coalition assigned to the $i^{th}$ task exceeds the required trait value for the $j^{th}$ trait. An element-wise max operation is performed between $E$ and the zero matrix. This removes all the values in $E$ which represent that an allocated coalition exceeds a required trait value and leaves the values that represent that a required trait value has not yet been met. An element-wise summation is then performed on the resulting matrix to compute the \textit{trait mismatch error}. This error is then normalized by the element-wise summation of the desired trait matrix \desiredTraitMatrix\ to compute the percentage trait mismatch error. When the error is zero, then the allocation satisfies the desired traits matrix.

APR does not use any information from the scheduling layer and, as such, tends to search the graph deeply. This is caused by nodes deeper in the graph having more robots assigned and a smaller desired trait mismatch error. This leads APR to find an allocation that satisfies the desired trait requirements quickly at the expense of ignoring schedules with considerably shorter makespan.

\subsubsection{Normalized Schedule Quality (NSQ)}
NSQ is a measure of how much an allocation minimizes the makespan of its accompanying schedule. Specifically, NSQ is calcuated as 
\begin{equation}
    \label{equ:nsq}
    f_{nsq}(C_{\overline{\schedule[bold]}}) = \frac{C_{\overline{\schedule[bold]}} -  C_{\schedule[bold]_{best}}}{C_{\schedule[bold]_{worst}} -  C_{\schedule[bold]_{best}}}    
\end{equation}
where $C_{\schedule}$ is the makespan, or completion time, of the schedule \schedule, $\overline{\schedule[bold]}$ is the schedule based on \taskNetwork[bold]\ and $\overline{\allocation[bold]}$,  $\schedule[bold]_{best}$ is the schedule without any constraints placed on the schedule from the allocation and motion planning, and $\schedule[bold]_{worst}$ is the schedule where the task network is totally-ordered and all motion plans are assumed to be the longest possible length.

The three variables $\bar{\schedule[bold]}$, $\schedule[bold]_{best}$, and $\schedule[bold]_{worst}$ are all computed by the scheduling layer. As NSQ only considers the schedule and not the allocation, it tends to favor a broader search. This is caused by nodes closer to the root having fewer constraints and, therefore, lower makespans. This leads to NSQ finding an allocation that satisfies the desired trait requirements with the minimum makespan at the expense of searching a much larger area of the graph.

\subsubsection{Time-Extended Task Allocation Quality (TETAQ)}
Using a convex combination of APR and NSQ, we create a heuristic that considers both the quality of the allocation as well as the quality of the schedule generated from it. TETAQ is calculated as
\begin{equation}
    \label{equ:tetaq}
    f_{tetaq} (\overline{\allocation}, C_{\overline{\schedule[bold]}}) = \alpha f_{apr}(\overline{\allocation}) + (1 - \alpha) f_{nsq}(C_{\overline{\schedule[bold]}})
\end{equation}
where $\alpha \in [0,1]$ is a user-specified parameter that controls each heuristic's relative influence. If $\alpha = 0$ then this heuristic is APR, and if $\alpha = 1$ then the heuristic is NSQ. TETAQ takes qualities from both to perform a search, which allows it to balance finding an allocation that satisfies the desired traits quickly with finding one that minimizes the makespan of the assigned robots' schedule.

\begin{algorithm}[tb!]
\DontPrintSemicolon % Some LaTeX compilers require you to use \dontprintsemicolon instead
\KwIn{$\angled{\taskNetwork[bold],\ \robotTraitMatrix[bold],\ \desiredTraitMatrix[bold],\ \setInitialConfigs[bold],\ \setConfigSpaces[bold], \alpha}$}
\KwOut{$\angled{\allocation[bold],\ \setMotionPlans[bold],\ \schedule[bold]}$}

root $\leftarrow$ empty allocation\;
pq $\leftarrow$ PriorityQueue(\{root\})\;
\While{pq is not empty}
{
    \customcomment{Identify the node with lowest tetaq value}\;
    node $\leftarrow$ pq.pop()\;
    
    \customcomment{Check if the node is a solution}\;
    \If{node.apr == 0\ \textbf{and} \ node.nsq $< \infty$}
    {
        \Return{node.\allocation,\ node.\setMotionPlans\ node.\schedule}
    }
    
    \customcomment{Compute heuristics for each successor}\;
    \For{child $\in$ generateSuccessors(node)}
    {
        \customcomment{Compute using \Cref{equ:apr}}\;
        child.apr $\leftarrow$ APR(child, \robotTraitMatrix[bold], \desiredTraitMatrix[bold])\;
        
        \customcomment{Compute schedule and motion plan}\;
        $\schedule[],\ \schedule[]_{best},\ \schedule[]_{worst},\ \setMotionPlans[]$ $\leftarrow$ schedule(child, \taskNetwork[bold])\;

        \customcomment{Compute using \Cref{equ:nsq}}\;
        child.nsq $\leftarrow$ NSQ($\schedule[],\ \schedule[]_{best},\ \schedule[]_{worst}$)\;

        \customcomment{Compute using \Cref{equ:tetaq}}\;
        child.tetaq = $\alpha\ * $ child.apr $+\ (1\ -\ \alpha)\ *$ child.nsq\;
        
        pq.push(child)\;
    }
}

\Return{Null}\;
\caption{ITAGS}
\label{algo:itags}
\end{algorithm}

\subsection{Scheduling and Motion Planning} 
The scheduling layer determines if it is feasible to schedule for a task network and an allocation and provides the makespan used by the heuristics. 
This layer builds the schedule based on three temporal components: the static duration of each task, the time needed for each robot to travel to the task's initial configuration, and the time needed for the assigned coalition of robots to execute the movements required to complete the task. 
To this end, the scheduling layer provides each task's initial and terminal configurations to the motion planning layer. 
%\cut{to determine if there is a feasible motion plan between the two configurations}. 

The motion planning layer uses the information received from the scheduling layer to determine if there is a feasible motion plan between the two configurations. If it can construct a motion plan, then the motion plan is sent back to the scheduling layer. To reduce computational costs and reuse motion plans, this layer memoizes the path for each specific pair of configurations and specific robot or coalition. 
% The memoization reduces the computational cost of generating motion plans and allows the motion planning layer to reuse the motion plans in subsequent queries.

The scheduling layer then uses the coalition's speed to determine the time needed to execute the motion plan. If a feasible schedule can be found then it calculates $\bar{\schedule[bold]}$, $\schedule[bold]_{best}$, and $\schedule[bold]_{worst}$. 

% %% Disjunctive Constraint
% \begin{definition}
% \label{def:disjunctive_constraint}
% A \textbf{Disjunctive constraint} is a relationship between two tasks, $t_a$ and $t_b$, that ensures that either $t_b$ must finish before $t_a$ starts or $t_a$ must finish before $t_b$ starts \cite{Bhargava2019MultiagentNetworksb}.
% \end{definition}

For a time-extended task allocation, a schedule has two different types of temporal constraints: precedence constraints and disjunctive constraints. The precedence constraints come from the task network. A \textbf{Disjunctive constraint} is a temporal relationship between two tasks, $t_a$ and $t_b$, that ensures that either $t_b$ must finish before $t_a$ starts or $t_a$ must finish before $t_b$ starts \cite{Bhargava2019MultiagentNetworksb}. A disjunctive constraint is created when a robot is assigned to a task $t_a$ making it unable to perform another task $t_b$ at the same time as $t_a$.

To find a schedule that adheres to both of these types of constraints, the scheduling layer uses a three-part scheduling approach. The first part converts the task network into a simple temporal network (STN) \cite{dechter1991temporal}, which provides a graphical representation of the start and end times for each task, where tasks are separated by precedence constraints. STNs are commonly used for scheduling due to their ability to be updated and checked for consistency in polynomial time \cite{Gini2017Multi-robotConstraints}.
% A STN provides a graphical representation of temporal events. Each vertex in the STN represents a time point, and each weighted directed edge represents inequality constraints between two time-points. When converting the task network to the STN, two vertices are created for each task, with one representing the start of the task and one representing the end of the task. For each task, an edge is added from the vertex representing its start time point to the vertex representing its end time point with the duration of the task as its weight. Additionally, each precedence constraint, $t_a \prec t_b$, is added as an edge between the vertex representing the end time point of $t_a$ to the vertex representing the start time point of $t_b$. 
We use a variant of the single-source shortest path algorithm \cite{dechter1991temporal} to compute the minimum makespan schedule from this STN. This schedule does not yet include the durations of any motion plans nor considers the disjunctive constraints imposed by the allocation, and so this schedule is $\schedule[bold]_{best}$ as used by NSQ. 

The second part of this approach adds in the disjunctive constraints from the allocation. In order to create a schedule, the disjunctive constraints need to be converted into precedence constraints by selecting one of their two orderings (i.e. for a disjunctive constraint between two tasks $t_a$ and $t_b$ selecting that either $t_a$ must finish before $t_b$ starts ($t_a \prec t_b$) or vise versa). We perform a tabu search\cite{Laguna2018TabuSearch} over the possible orderings for each disjunctive constraint while minimizing the makespan of the schedule. Each node in this search is the STN for $\schedule[bold]_{best}$ with additional edges added for the precedence constraints created from the ordering selection of the disjunctive constraints. The resulting STN from this part of the approach is called STN$_d$.

The third part of this approach adds in the execution durations of the motion plans to the STN. During this step, the motion planning layer determines if every required motion plan is feasible. If feasible, the length of each motion plan is used by the scheduling layer to determine the execution duration of the motion plan.

If it is not feasible to find a schedule for an allocation, then the task allocation layer is alerted, and the allocation is pruned from the search. Finding a schedule is infeasible if:
\begin{flushitemize}
    \item the STN for $\schedule[bold]_{best}$ is temporally inconsistent
    \item no temporally consistent STN can be found during the tabu search
    \item the motion planner times out
    %\item \cut{the addition of a motion plan causes the STN$_d$ to become temporally inconsistent}
\end{flushitemize}

Finally, we compute the makespan of $\schedule[bold]_{worst}$. For computational efficiency, we approximate an over-estimation of the makespan of the worst possible schedule without having any robots slow down or wait,
$$C_{\schedule[bold]_{worst}} = \frac{2 M z}{w} + \sum_{m=1}^M dur(t_m)$$
where \numTasks\ is the number of tasks in \taskNetwork, $dur(t_m)$ is the static duration of task $t_m$, $z$ is the length of the longest possible path in \setConfigSpaces, and $w$ is the speed of the slowest robot.

\section{Evaluation}

\begin{figure*}[t!]
    \centering
    \includegraphics[width=0.75\textwidth]{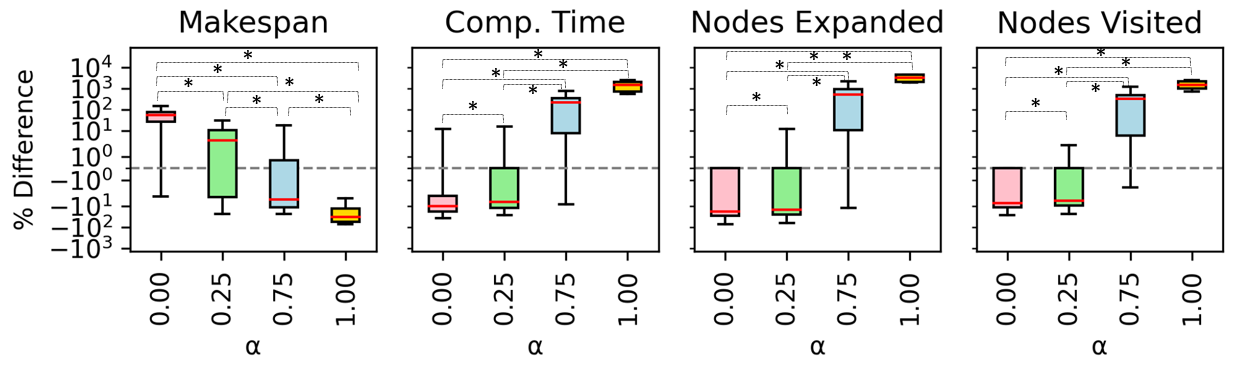}
    \caption{The results of ITAGS with various $\alpha$ values normalized with respect to ITAGS$_{\alpha=0.5}$. $y=0$ corresponds to ITAGS$_{\alpha=0.5}$. Anything above $y=0$ is worse than ITAGS$_{\alpha=0.5}$ and conversely anything below is better. `*' denotes statistical significance with a p-value $<$ 0.05.}
    \label{fig:alphas}
\end{figure*}

We evaluated ITAGS using three sets of experiments in a simulated emergency response domain~\cite{kitano1999, bechon2014hipop, Messing2020ForwardPlanning, Whitbrook2015ASystems, Zhao2016AScenario}. In this domain, a diverse set of robots with different traits need to work together to rescue wounded survivors, deliver medicine to hospitals, put out fires, and rebuild damaged infrastructure.

For the first two experiments, we generated a set of $105$ problems from this domain by randomly sampling the number of robots, survivors, fires, and damaged buildings. Each problem had between 6-12 robots and 12-45 tasks. We also randomized the locations of the survivors, fires, damaged buildings, hospitals, and the robots' initial location. In the first set of experiments, we evaluated the effectiveness and the relative influence of our heuristics. In the second set of experiments, we evaluated the effectiveness of ITAGS's interleaved allocation and scheduling by comparing it to a baseline that uses a sequential version of ITAGS. For both experiments, we report computation time, the number of nodes expanded and visited, and the makespan of the generated schedule as metrics.

For the third experiment, we generated a set of $50$ problems from this domain. Each problem had 20 robots and 40 tasks. The problems were randomized similarly to the problems for the first two experiments. For this experiment, we compared ITAGS against two state-of-the-art ST-MR-TA task allocation algorithms: CFLA2 and CCF\cite{Capezzuto2020}. These algorithms represent the most recent efforts in solving the ST-MR-TA task allocation problem.

% \begin{figure}[!b]
%     \centering
%     \includegraphics[width=0.6\columnwidth]{Images/map.png}
%     \caption{Example map. Buildings are in black and roads are gray.}
%     \figlabel{map}
% \end{figure}

For all of the experiments, maps from the Robocup Rescue Competition \cite{kitano1999} were used. 
% Each map contains buildings of various polygon shapes and roads. 
% An example map is shown in \figref{map}. 
For the motion planning layer, we used a Lazy PRM \cite{Bohlin2000PathPRM} implementation from the Open Motion Planning Library \cite{Sucan2012TheLibrary}. We ran all experiments on an i7-8565 CPU with 16GB of RAM.

\subsection{Relative Influence of APR and NSQ on Performance}

The first experiment involved ablation studies to investigate the relative influence of APR and NSQ on performance. To this end, we generated five variants of ITAGS, each with different values for $\alpha$ ($0,\ 0.25,\ 0.5,\ 0.75,$ and $1$) from Equation (\ref{equ:tetaq}). Note that these values for $\alpha$ indicate different relative weightings of NSQ and APR.
% In the first experiment, we compare three variants of TETAQ, each with a different $\alpha$ (where $\alpha$ is a user-specified parameter that controls the relative influence of each heuristic) value, against NSQ and APR. 
The results of these tests can be seen in \Cref{fig:alphas}. 
To make the relationship between the weighting of the heuristics and the performance clear, \Cref{fig:alphas} displays the results normalized to an equal weighting of both heuristics ($\alpha=0.5$). As such, the horizontal line for $y=0$ represents the results of $\alpha=0.5$. Anything above $y=0$ represents a result that was larger than the $\alpha=0.5$ baseline (e.g. took more time to compute or expanded more nodes), and anything below $y=0$ represents a result that was smaller than the $\alpha=0.5$ baseline (e.g. took less time to compute or expanded fewer nodes).

All $\alpha$ values were capable of solving all 105 problems except $\alpha=1.0$, which only solved 7.5\% of the problems. As $\alpha \rightarrow 1$, it causes the search to focus more on minimizing the makespan, thereby mimicking a breadth-first search. This is because adding more robots increases the number of constraints and, subsequently, the makespan. As a result of these broader searches, ITAGS$_{\alpha=1}$ is more likely to run out of memory and fail to solve the problem.

We performed a Kruskal-Wallis test followed by a post-hoc Dunn's test to show the statistical significance of each of the tested $\alpha$ values. These tests' results are shown in \Cref{fig:alphas} with `*' denoting p-values $<$ 0.5. These results indicate that $\alpha$ changes have a statistically significant effect on the computation time, makespan, the number of nodes expanded, and the number of nodes explored.

We find that as $\alpha$ increased, computation time increased, and makespan decreased. Specifically, as $\alpha \rightarrow 1$, the search tends to produce solutions with lower makespans. However, it also tends to require higher computation times as a result of visiting and exploring more nodes. Conversely, as $\alpha \rightarrow 0$, the search tends to visit and explore fewer nodes, leading to lower computation times. However, this comes at the expense of a longer makespan. These experiments show that there is a balance between minimizing makespan and minimizing computation time. 
Our combined heuristic TETAQ merges the effects of the two heuristics to balance both computational efficiency and solution quality.

\subsection{Effects of Interleaving on Performance}

\begin{figure}[!b]
    \centering
    \includegraphics[width=0.7\columnwidth]{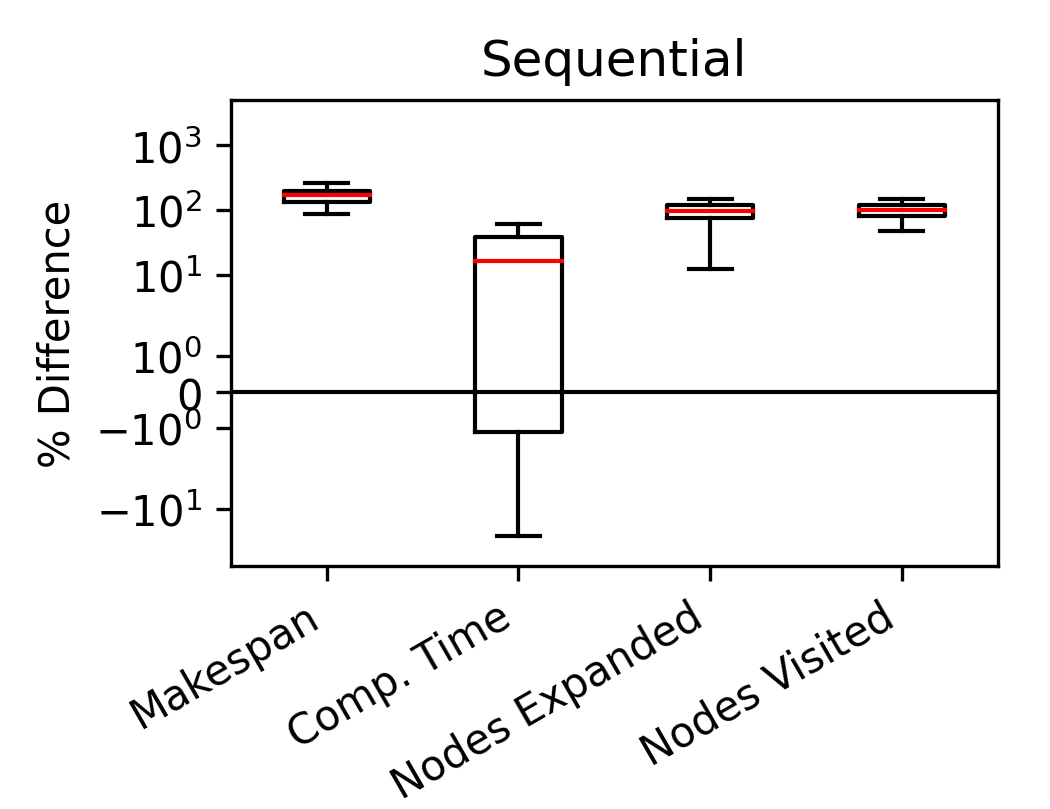}
    \caption{The results of the sequential version of ITAGS (ITAGS$_S$) normalized with respect to ITAGS$_{\alpha=0.5}$. $y=0$ represents ITAGS$_{\alpha=0.5}$. Anything above $y=0$ is worse than ITAGS$_{\alpha=0.5}$ and conversely anything below is better.}
    \label{fig:sequential}
\end{figure}

In this experiment, we created a sequential version of ITAGS, known as ITAGS$_S$. Instead of using scheduling and motion planning to guide the search for task allocation, ITAGS$_S$ completes each operation in sequence. It searches the allocation graph until an allocation that satisfies the trait requirements is found. If possible, ITAGS$_S$ schedules the found allocation and updates the schedule with the execution times of motion plans. If either a schedule cannot be created because the allocation is temporally inconsistent or one of the motion planning queries is infeasible,
then ITAGS$_S$ continues the search until a satisficing allocation is found. 

The results of the comparison can be found in \Cref{fig:sequential}. Similar to the first experiment, we normalized the metrics with respect to ITAGS$_{\alpha=0.5}$. As can be seen, ITAGS$_{\alpha=0.5}$ consistently outperformed the sequential version ITAGS$_S$ in terms of all of the metrics. On average, ITAGS$_S$ generated an output with a 168\% longer makespan while taking 17\% longer to compute a valid result. It also visited 99\% more nodes and explored 97\% more nodes on average. 

There are two notable reasons for ITAG's superior performance. First, ITAGS uses its heuristics to prune partial allocations for which scheduling and motion planning are infeasible. In contrast, ITAGS$_S$ is likely to end up exploring branches in the allocation graph that violate scheduling or motion planning constraints. Second, ITAGS also minimizes the makespan of the schedule. However, ITAGS$_S$ does not consider the schedule while searching for an allocation. This leads ITAGS$_S$ to prefer allocating most tasks to robots with higher trait values. As such, ITAGS$_S$ results in a reduced number of concurrent tasks.

\subsection{Comparison against CFLA2 and CCF}

In the third experiment, we compared ITAGS against two state-of-the-art ST-MR-TA task allocation algorithms in CFLA2 and CCF\cite{Capezzuto2020}. Both of these algorithms operate on a more-restricted problem structure that does not involve trait-based agent/task modeling, ordering constraints, or non-graph-based motion planning. Thus in order to benchmark against these algorithms a preprocessing stage was added. 

For each problem, the preprocessing stage first generates a fully connected graph where each node is a location (i.e. the hospital, survivor 1's initial location, etc.) and each edge is labeled with the time required to traverse for each robot/coalition. In order to calculate the traversal times, the motion planning module from ITAGS is queried for motion plans between each pair of nodes. The length of each motion plan and the speed of each robot/coalition is then used to compute the traversal times for each edge. 

Next, the preprocessing stage accommodates for the fact that the baselines cannot handle trait-based models. The baselines model each task $t$ as having a certain amount of work that needs to be accomplished ($w_t$). They also have a utility function $u(t, c)$ which computes how fast a specific coalition $c$ can work on a specific task $t$. The duration of a task $t$ can be computed as $dur(t, c) = w_t / u(t, c)$. The preprocessing stage sets $w_t$ to a constant value for all tasks. It then creates the utility function $u(t, c)$ such that the $dur(t, c)$ is the same as in the problem description for all tasks. If a coalition's collective traits do not satisfy the task's requirements then the utility function returns zero.

\begin{figure}[!t]
    \centering
    \begin{subfigure}[t]{1\columnwidth}
        \includegraphics[width=\textwidth]{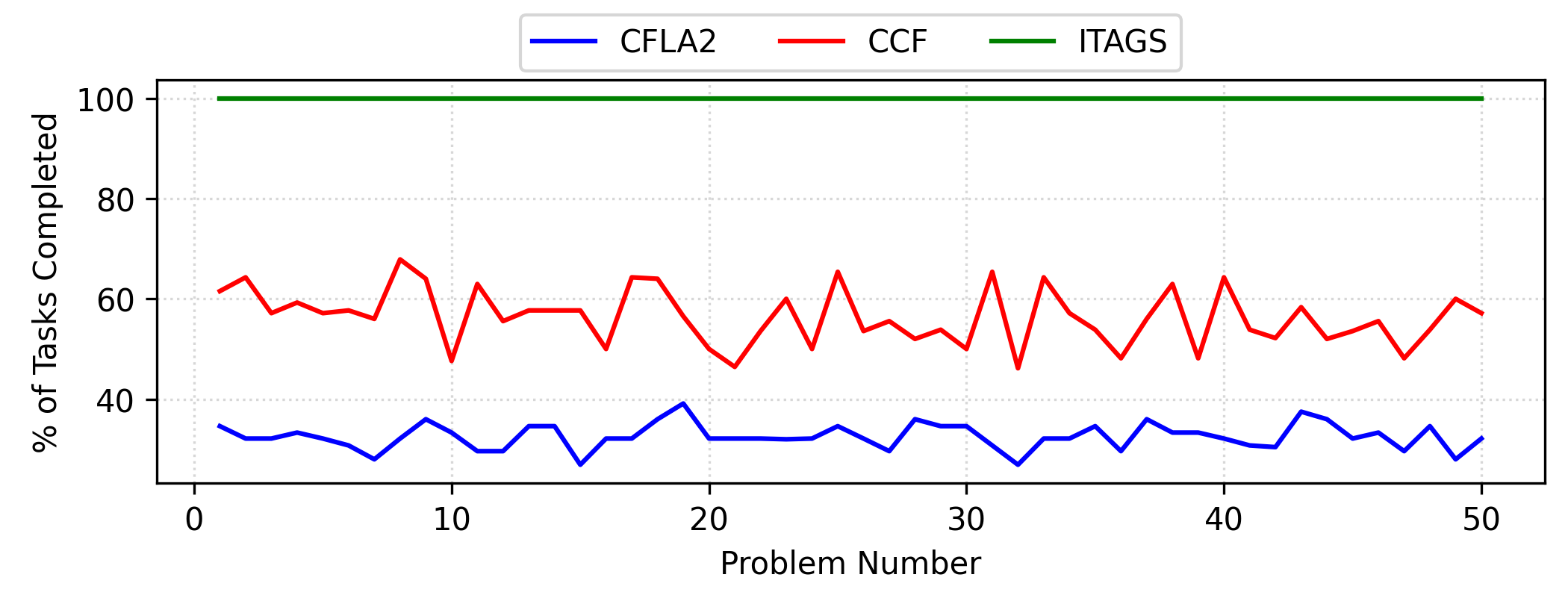}
        \caption{The percentage of tasks that each algorithm was able to allocate for each problem.}
        \figlabel{task_completed}
    \end{subfigure}
    
    \begin{subfigure}[t]{1\columnwidth}
        \includegraphics[width=\textwidth]{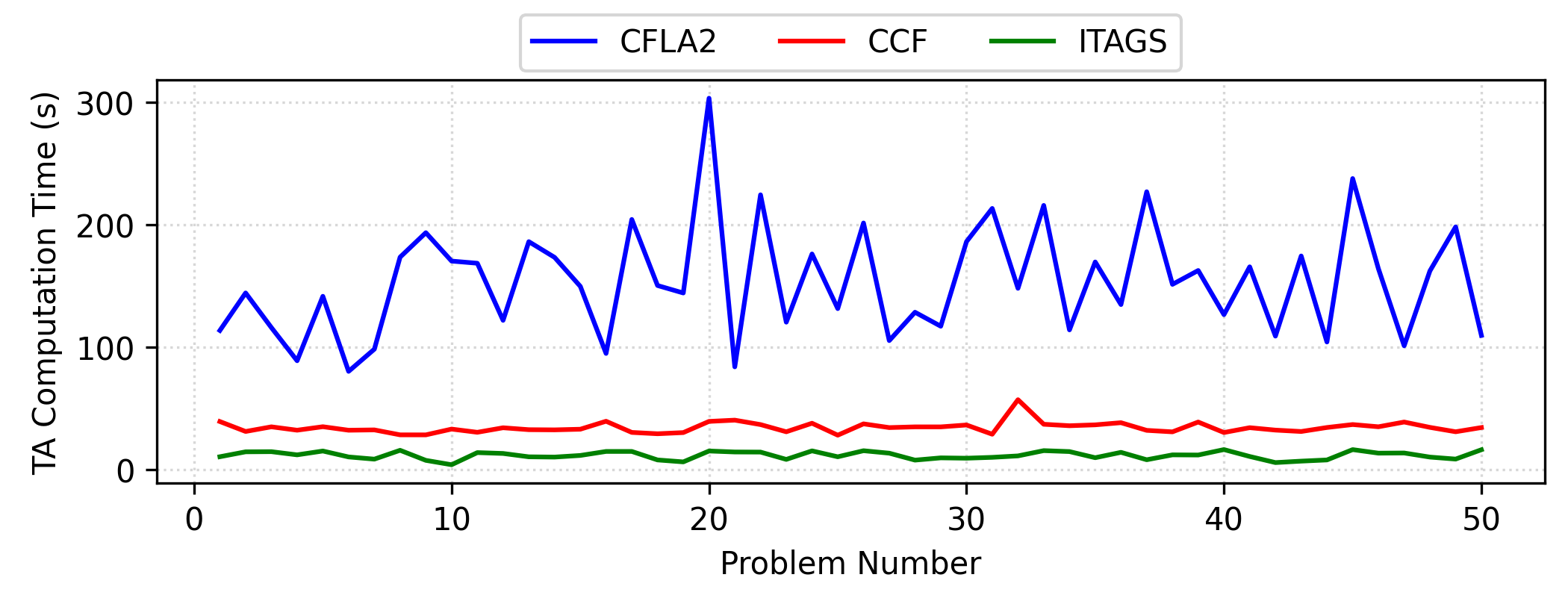}
        \caption{The amount of time spent by each algorithm on everything except motion planning for each problem.}
        \figlabel{ta_comp_time}
    \end{subfigure}
    
    % \begin{subfigure}[t]{1.0\columnwidth}
    %     \includegraphics[width=\textwidth]{Images/comp_time_1000.png}
    %     \caption{The total time spent by each algorithm for each problem.}
    %     \figlabel{comp_time}
    % \end{subfigure}
    
    \caption{Benchmark against CFLA2 and CCF }
\end{figure}

As shown in \figref{task_completed}, ITAGS is able to allocate and schedule all tasks for each of the problems. This observation is explained by the fact that, unlike CFLA2 and CCF, ITAGS requires allocating and scheduling of all tasks. Indeed, both baselines fail to fully allocate and schedule any of the problems, with CFLA2 and CCF averaging 32.4\% and 56.4\% of tasks allocated and scheduled, respectively. We observe that baselines sometimes allocate agents to a task without accounting for the task's trait requirements. This leads to incomplete tasks as robots can get stuck on certain tasks without contributing to the tasks' completion. Additionally, without deadlines, CFLA2's look-ahead phase effectively results in a random choice and impedes its ability to allocate robots to tasks. CCF assumes that every assignment improves task progress irrespective of the current set of robots assigned to the task. However, this assumption does not hold for tasks in which robots cannot make progress without crossing a minimum threshold in terms of aggregated traits. For instance, consider a task that involves moving a 10 kg object. Assigning a single robot with a 5 kg payload will not contribute to 50\% task completion.

\figref{ta_comp_time} shows the total computation time excluding the time needed to generate the fully connected graph for CFLA2 and CCF and the time spent motion planning for ITAGS. This ensures that we are comparing only the task allocation and scheduling capabilities of the algorithms. As seen, ITAGS allocates and schedules all tasks while taking less time to compute a solution. On average, ITAGS spends 11.54s on task allocation and scheduling when computing a solution, while CFLA2 and CCF spend 153.61s and 34.27s, respectively. There are a few reasons that ITAGS performs better than both CFLA2 and CFF. First, ITAGS only considers the start and end time-points for each task when scheduling, whereas both CFLA2 and CCF step through time with discrete timesteps and greedily allocate robots to tasks. Second, ITAGS focuses on simultaneously minimizing the makespan and trait mismatch error.  Being able to consider the traits allows ITAGS to model the relationship between agents and task better, leading to more efficient searching of possible allocations. On the other hand, CFLA2 and CCF focus on allocating as many tasks at a time as possible without considering whether the robot's traits contribute to satisfying the trait requirements of a task. CCF specifically also prioritizes allocating a robot to tasks such that travel time is minimized.
% This can lead to both CFLA2 and CCF allocating robots to tasks that the robot cannot contribute. 
Third, ITAGS considers the entire schedule as it solves the problem. This allows ITAGS to consider actions throughout the schedule and how they affect the overall makespan. CCF only considers a single timestep at a time, and as a result some of its allocations inadvertently create a bottleneck for future allocations. CFLA2 does have a look-ahead process, but with no deadlines, it effectively results in a random choice. Furthermore, it creates considerable computational burden for this result. The look-ahead process and making allocations for only a single timestep have a detrimental impact on computation time.

\subsection{Summary}

The first experiment empirically demonstrates the trade-offs involved in prioritizing either \textit{Allocation Percentage Remaining} or \textit{Normalized Schedule Quality}. 
Results from the second experiment indicate that the hierarchical interleaved approach of ITAGS consistently outperforms a baseline approach that sequentially performs allocation, scheduling, and motion planning.
In the third experiment, ITAGS is shown to empirically outperform state-of-the-art ST-MR-TA task allocation algorithms CFLA2 and CCF \cite{Capezzuto2020}.

\section{Conclusions}

%Heterogeneous multi-robot systems offer the ability to solve problems that would be infeasible for a single robot. The cooperation of robots and the leveraging of individual robots' relative strengths is key to the strength of these multi-robot systems. For this reason, algorithms that support the allocation and cooperation of multi-robot teams are needed.

We introduced a unified framework that interleaves task allocation, scheduling, and motion planning for heterogeneous multi-robot systems. We also presented the iterative search method for solving the trait-based time-extend task allocation problem. To guide the search and enable interleaving, we developed two heuristics, one based on the quality of allocation and another based on the time needed to execute the associated schedule and motion plans. Further, we empirically demonstrated the trade-offs involved in choosing the relative weighting of the two heuristics. 
Our experiments in a simulated emergency response domain conclusively demonstrate the effectiveness and the relative advantages of our interleaved approach over a sequential baseline and two state-of-the-art approaches.
% This interleaved system allows for communication between the task allocation and scheduling layers to allow each of these systems to operate more effectively. 
% Through our experiments, we were able to show that the interleaving used by our framework is capable of outperforming similar frameworks that work sequentially and do not allow for communication between their components. 

%These heuristics provide a balanced approach to task allocation with APR quickly finding an allocation that meets the task requirement at the expense of the schedule's makespan and NSQ focusing at improving the schedules makespan at the expense of computation time. 
% We were able to show that combining these heuristics can be used to vary these tradeoffs using TETAQ. 

%\input{sections/9_FutureWork}

% \input{sections/10_acknowledgements} 

\bibliographystyle{ieeetr}
\bibliography{references}

\begin{thebibliography}{10}

\bibitem{Roldan2016HeterogeneousGreenhouses}
J.~J. Rold{\'{a}}n, P.~Garcia-Aunon, M.~Garz{\'{o}}n, J.~de~Le{\'{o}}n, J.~del
  Cerro, and A.~Barrientos, ``{Heterogeneous multi-robot system for mapping
  environmental variables of greenhouses},'' {\em Sensors}, vol.~16, no.~7,
  2016.

\bibitem{McCook2007a}
C.~McCook and J.~Esposito, ``{Flocking for Heterogeneous Robot Swarms: A
  Military Convoy Scenario},'' in {\em Southeastern Symposium on System
  Theory}, pp.~26 -- 31, 2007.

\bibitem{Stroupe2005HeterogeneousServicing}
A.~W. Stroupe, T.~Huntsberger, B.~Kennedy, H.~Aghazarian, E.~T. Baumgartner,
  A.~Ganino, M.~Garrett, A.~Okon, M.~Robinson, and J.~A. Townsend,
  ``{Heterogeneous robotic systems for assembly and servicing},'' in {\em
  International Symposium on Artifical Intelligence, Robotics and Automation in
  Space}, 2005.

\bibitem{Baras2019AnWarehouses}
N.~Baras and M.~Dasygenis, ``{An algorithm for routing heterogeneous vehicles
  in robotized warehouses},'' in {\em Panhellenic Conference on Electronics and
  Telecommunications}, 2019.

\bibitem{Gerkey2004ASystems}
B.~P. Gerkey and M.~J. Matari{\'{c}}, ``{A formal analysis and taxonomy of task
  allocation in multi-robot systems},'' {\em International Journal of Robotics
  Research}, vol.~23, no.~9, pp.~939--954, 2004.

\bibitem{Korsah2013}
G.~A. Korsah, M.~B. Dias, and A.~Stentz, ``{A Comprehensive Taxonomy for
  Multi-Robot Task Allocation},'' {\em International Journal of Robotics
  Research}, vol.~32, no.~12, pp.~1495--1512, 2013.

\bibitem{Ravichandar2019}
H.~Ravichandar, K.~Shaw, and S.~Chernova, ``{STRATA: A Unified Framework for
  Task Assignments in Large Teams of Heterogeneous Agents},'' {\em Journal of
  Autonomous Agents and Multi-Agent Systems}, 2019.

\bibitem{Prorok2017TheSwarms}
A.~Prorok, M.~A. Hsieh, and V.~Kumar, ``{The Impact of Diversity on Optimal
  Control Policies for Heterogeneous Robot Swarms},'' {\em IEEE Transactions on
  Robotics}, 2017.

\bibitem{Gerkey2003Multi-robotArchitectures}
B.~P. Gerkey and M.~J. Matari{\'{c}}, ``{Multi-robot task allocation: Analyzing
  the complexity and optimality of key architectures},'' in {\em International
  Conference on Robotics and Automation}, vol.~3, 2003.

\bibitem{Nunes2017}
E.~Nunes, M.~Manner, H.~Mitiche, and M.~Gini, ``{A taxonomy for task allocation
  problems with temporal and ordering constraints},'' {\em Robotics and
  Autonomous Systems}, vol.~90, 2017.

\bibitem{Sariel2006}
S.~Sariel and T.~Balch, ``{A distributed multi-robot cooperation framework for
  real time task achievement},'' {\em Distributed Autonomous Robotic Systems},
  pp.~187--196, 2006.

\bibitem{Jones2011Time-extendedConstraints}
E.~G. Jones, M.~B. Dias, and A.~Stentz, ``{Time-extended multi-robot
  coordination for domains with intra-path constraints},'' {\em Autonomous
  Robots}, vol.~30, pp.~41--56, 2011.

\bibitem{Giordani2013}
S.~Giordani, M.~Lujak, and F.~Martinelli, ``{A distributed multi-agent
  production planning and scheduling framework for mobile robots},'' {\em
  Computers and Industrial Engineering}, vol.~64, no.~1, pp.~19--30, 2013.

\bibitem{Krizmancic2020}
M.~Krizmancic, B.~Arbanas, T.~Petrovic, F.~Petric, and S.~Bogdan,
  ``{Cooperative Aerial-Ground Multi-Robot System for Automated Construction
  Tasks},'' {\em IEEE Robotics and Automation Letters}, vol.~5, no.~2,
  pp.~798--805, 2020.

\bibitem{Tompkins2003OptimizationOperationsb}
M.~F. Tompkins, {\em {Optimization Techniques for Task Allocation and
  Scheduling in Distributed Multi-Agent Operations}}.
\newblock PhD thesis, Massachusetts Institute of Technology, 2003.

\bibitem{Korsah2012}
G.~A. Korsah, B.~Kannan, B.~Browning, A.~Stentz, and M.~B. Dias, ``{xBots: An
  approach to generating and executing optimal multi-robot plans with
  cross-schedule dependencies},'' {\em International Conference on Robotics and
  Automation}, pp.~115--122, 2012.

\bibitem{Guerrero2017Multi-robotSolutions}
J.~Guerrero, G.~Oliver, and O.~Valero, ``{Multi-robot coalitions formation with
  deadlines: Complexity analysis and solutions},'' {\em PLoS ONE}, vol.~12,
  no.~1, 2017.

\bibitem{Thakar2019TaskManipulation}
S.~Thakar, A.~Kabir, P.~M. Bhatt, R.~K. Malhan, P.~Rajendran, B.~C. Shah, and
  S.~K. Gupta, ``{Task assignment and motion planning for bi-manual mobile
  manipulation},'' in {\em International Conference on Automation Science and
  Engineering}, 2019.

\bibitem{Kabir2020IncorporatingManipulators}
A.~M. Kabir, S.~Thakar, P.~M. Bhatt, R.~K. Malhan, P.~Rajendran, B.~C. Shah,
  and S.~K. Gupta, ``{Incorporating Motion Planning Feasibility Considerations
  during Task-Agent Assignment to Perform Complex Tasks Using Mobile
  Manipulators},'' in {\em International Conference on Robotics and
  Automation}, 2020.

\bibitem{Ramchurn2010CoalitionConstraints}
S.~D. Ramchurn, M.~Polukarov, A.~Farinelli, C.~Truong, and N.~R. Jennings,
  ``{Coalition formation with spatial and temporal constraints},'' in {\em
  International Joint Conference on Autonomous Agents and Multiagent Systems},
  2010.

\bibitem{Capezzuto2020}
L.~Capezzuto, D.~Tarapore, and S.~D. Ramchurn, ``{Anytime and Efficient
  Coalition Formation with Spatial and Temporal Constraints},'' in {\em
  European Conference on Multi-Agent Systems}, 2020.

\bibitem{Koes2005}
M.~Koes, I.~Nourbakhsh, and K.~Sycara, ``{Heterogeneous multirobot coordination
  with spatial and temporal constraints},'' {\em AAAI Workshop - Technical
  Report}, vol.~WS-05-06, pp.~9--16, 2005.

\bibitem{Weld1994AnPlanning}
D.~S. Weld, ``{An Introduction to Least Commitment Planning},'' {\em AI
  magazine}, vol.~15, no.~4, pp.~27--27, 1994.

\bibitem{Choset2005PrinciplesImplementations}
H.~Choset, K.~Lynch, S.~Hutchinson, G.~Kantor, W.~Burgard, L.~Kavraki, and
  S.~Thrun, {\em {Principles of Robot Motion: Theory, Algorithsm, and
  Implementations}}.
\newblock MIT Press, 2005.

\bibitem{Bhargava2019MultiagentNetworksb}
N.~Bhargava and B.~Williams, ``{Multiagent disjunctive temporal networks},'' in
  {\em Proceedings of the International Joint Conference on Autonomous Agents
  and Multiagent Systems, AAMAS}, vol.~1, 2019.

\bibitem{dechter1991temporal}
R.~Dechter, I.~Meiri, and J.~Pearl, ``{Temporal constraint networks},'' {\em
  Artificial Intelligence}, vol.~49, no.~1-3, pp.~61--95, 1991.

\bibitem{Gini2017Multi-robotConstraints}
M.~Gini, ``{Multi-robot allocation of tasks with temporal and ordering
  constraints},'' in {\em AAAI Conference on Artificial Intelligence}, 2017.

\bibitem{Laguna2018TabuSearch}
M.~Laguna, ``{Tabu search},'' in {\em Handbook of Heuristics}, vol.~1-2, 2018.

\bibitem{kitano1999}
H.~Kitano, S.~Tadokoro, I.~Noda, H.~Matsubara, T.~Takahashi, A.~Shinjou, and
  S.~Shimada, ``{Robocup rescue: Search and rescue in large-scale disasters as
  a domain for autonomous agents research},'' in {\em International Conference
  on Systems, Man, and Cybernetics}, 1999.

\bibitem{bechon2014hipop}
P.~Bechon, M.~Barbier, G.~Infantes, C.~Lesire, and V.~Vidal, ``{HiPOP:
  Hierarchical Partial-Order Planning},'' {\em STAIRS}, pp.~51--60, 2014.

\bibitem{Messing2020ForwardPlanning}
A.~Messing and S.~Hutchinson, ``{Forward Chaining Hierarchical Partial-Order
  Planning},'' in {\em The Workshop on the Algorithmic Foundations of Robotics
  (WAFR)}, 2020.

\bibitem{Whitbrook2015ASystems}
A.~Whitbrook, Q.~Meng, and P.~Chung, ``{A Novel Distributed Scheduling
  Algorithm for Time-Critical MultiAgent Systems},'' in {\em International
  Conference on Intelligent Robots and Systems}, 2015.

\bibitem{Zhao2016AScenario}
W.~Zhao, Q.~Meng, and P.~W. Chung, ``{A Heuristic Distributed Task Allocation
  Method for Multivehicle Multitask Problems and Its Application to Search and
  Rescue Scenario},'' {\em IEEE Transactions on Cybernetics}, vol.~46,
  pp.~902--915, 4 2016.

\bibitem{Bohlin2000PathPRM}
R.~Bohlin and L.~Kavraki, ``{Path Planning Using Lazy PRM},'' in {\em
  International Conference on Robotics and Automation (ICRA)}, 2000.

\bibitem{Sucan2012TheLibrary}
I.~Sucan, M.~Moll, and L.~Kavraki, ``{The Open Motion Planning Library},'' {\em
  IEEE Robotics {\&} Automation Magazine}, 2012.

\end{thebibliography}

\end{document}